\documentclass[a4paper]{article} 

\usepackage{natbib}

\usepackage[english]{babel}
\usepackage[utf8x]{inputenc}
\usepackage[T1]{fontenc}

\usepackage{amsmath}
\usepackage{amsfonts}
\usepackage{amssymb}

\usepackage{authblk}

\usepackage{graphicx}
\usepackage{diagbox}
\usepackage{hyperref}
\usepackage[export]{adjustbox}  
\usepackage{siunitx}  
\usepackage{booktabs} 
\usepackage[linesnumbered,ruled,vlined,]{algorithm2e}

\SetCommentSty{mycommfont}
\usepackage{tikz}
\usetikzlibrary{arrows}
\usepackage{url}

\newcommand{\eq}{\leftarrow}
\newcommand{\ie}{\emph{i.e.}}
\newcommand{\eg}{\emph{e.g.}}
\newcommand{\dx}{{\bar{x}}}	
\newcommand{\dy}{{\bar{y}}}
\newcommand{\dz}{{\bar{z}}}
\newcommand{\tx}{{\dot{x}}}	
\newcommand{\ty}{{\dot{y}}}

\newcommand{\sx}{{\tilde{x}}}	
\newcommand{\sy}{{\tilde{y}}}
\newcommand{\sz}{{\tilde{z}}}

\begin{document}

\title{The Sea Exploration Problem \\ \large 
  Data-driven Orienteering on a Continuous Surface}

\author[1]{Jo\~ao Pedro Pedroso\thanks{jpp@fc.up.pt}}
\author[1]{Alpar Vajk Kramer\thanks{akramer@inesctec.pt}}
\author[2]{Ke Zhang\thanks{kzhang@whut.edu.cn}}
\affil[1]{INESC TEC and Faculdade de Ci\^encias, Universidade do Porto, Portugal}
\affil[2]{School of Information Engineering, Wuhan University of Technology, China}

\providecommand{\keywords}[1]{\textbf{\textit{Keywords}} #1}

\date{April 2019}

\maketitle

\begin{abstract}
  This paper describes a problem arising in sea exploration, where the aim is to schedule the expedition of a ship for collecting information about the resources on the seafloor.  The aim is to collect data by probing on a set of carefully chosen locations, so that the information available is optimally enriched.  This problem has similarities with the orienteering problem, where the aim is to plan a time-limited trip for visiting a set of vertices, collecting a prize at each of them, in such a way that the total value collected is maximum.  In our problem, the score at each vertex is associated with an estimation of the level of the resource on the given surface, which is done by regression using Gaussian processes.  Hence, there is a correlation among scores on the selected vertices; this is a first difference with respect to the standard orienteering problem.  The second difference is the location of each vertex, which in our problem is a freely chosen point on a given surface.
\end{abstract}


\section{Introduction}
\label{sec:introd}

Sea exploration is important for countries with large areas in the ocean under their control, since in the future it may be possible to exploit some of the resources in the seafloor; for example, in the ``Solwara 1 Project\footnote{\url{https://web.archive.org/web/20100812102807/http://www.nautilusminerals.com/s/Projects-Solwara.asp}}'', mining of high grade copper and gold in the waters off Papua New Guinea has been considered.  However, seafloor contents are largely unknown; for characterizing them, a preliminary step is to fetch information about its composition.  This is currently being done by sending a ship in an expedition during which an underwater robot, or other equipment, collects samples at selected points.  Such expeditions are typically very costly; additionally, the ship must be available for other commitments at a predetermined port within a rigid and tight time limit.
{ 
Due to its costs, expeditions are relatively rare; in the applications that we are aware of, there is typically a trip to a given area in the period of several years\footnote{See, \eg, \url{https://www.iodp.org/expeditions/expeditions-schedule}}.
Nowadays, planning is usually done by experts, based on previously collected information and on intuition; because of the importance of the trips for inventorying seafloor resources, a method for helping decision makers carrying out the ship's schedule are desirable.  The aim of this paper is to provide a step in this path.  To our knowledge, this is the first attempt to address this problem.
}
Even though this paper describes the problem in the context of sea exploration, similar problems arise in other contexts (\eg, fire detection by drones on a forest).

More formally, the aim is to schedule the journey of a ship for collecting information about the resources of the seafloor (\eg, composition in certain materials).  The surface being considered here represented as a given (bounded) surface~$S\subset\mathbb{R}^2$.  For the sake of simplicity, we consider that the actual resource level at any point $(x,y) \in S$ can be conveyed by a real number, denoted by~$v(x,y)$.  This \emph{true value} is unknown, except for a limited number $N$ of points in~$S$ for which there is previous empirical information.

Optimal expedition planning involves three subproblems, each corresponding to a different phase on the process.
This first is \textbf{assessment}, which consists of the following: given a finite set of points for which the contents are known, build and indicator function $h(x,y)$ that associates to each point $(x,y) \in S$ the ``attractiveness'' (a real number) for exploring it, in terms of information that can be gathered in case that point is selected for probing.  
The second subproblem is \textbf{planning}, \ie, deciding on the position of a certain number $n$ of points to probe in the next expedition so as to maximize the overall informational reward; the duration of the trip 
{ 
  includes time for probing the chosen points and traveling between them, and is limited to a known bound which implicitly limits~$n$. 
}
The third subproblem, \textbf{estimation}, is related to the final aim of the problem, which is to have an evaluation $w(x,y)$ of the resource level available at any point on the surface~$S$, based on all the information available at the end of the trip.

This paper is organized as follows.  In Section~\ref{sec:litreview} we put this problem in the context of the literature.  Section~\ref{sec:method} describes the method that we developed of tackling the problem; results obtained with it are presented in Section~\ref{sec:results}.  Section~\ref{sec:conclusions} presents some conclusions and future research directions.

\section{Related work}
\label{sec:litreview}

{ 
  Our problem involves multiple research areas.   Subproblem 2, planning, is studied in artificial intelligence and operational research; see, \eg, \citet{ghallab2016automated} for planning in the setting of artificial intelligence and \citet{archetti2014chapter} for related integer optimization models.  Subproblems 1 and 3, assessment and estimation, belong to the areas of machine learning and data mining (see, \eg, \citet{robert2014machine} and \citet{han2011data}).  In particular, our problem's context can be seen as a particular case of active learning, where arbitrary points on the relevant surface can be selected for probing in such a way that the pattern to be discovered---resource content values allover the surface---is optimally enhanced.  Typical applications of active learning for data acquisition involve a discrete set of entities, for which additional information may be gathered if required.  For example, in \citet{zheng2002active} models for customer behavior are built, and data is acquired for selected customers in order improve the corresponding model.  Mentioned examples are the classification of customers into those who do transactions or not; the data acquisition problem is determining how many and which customers from which data should be acquired.  As our space for sampling is continuous and the model that we want to build involves all the samples, our setting is substantially different.  It is closer to active learning via query synthesis \citep{wang2015active},  where, in a classification context, the focus is on the synthesis of (unlabeled) instances close to the decision boundary, which are chosen for annotation, so that at the end a small labeled set can be used without compromising the solution quality.  In the regression context, the method presented in \citet{demir2014multiple} selects the most informative unlabeled samples for expanding a small initial training set, in the context of $\epsilon$-insensitive support vector regression.  This method is based on the evaluation of three criteria for the selection of samples to be labeled: relevancy, diversity and density.  A two step procedure based on clustering identifies the most relevant unlabeled samples, assuming that training samples in the same cluster of support vectors are the most relevant; among these, the most ``diverse'' samples---lying on different clusters in another clustering process---are chosen.  As this method relies on clustering of unlabeled samples, it is not obvious how it could be adapted to our setting.

  This section presents a literature review for each of the subproblems involved, assessment, planning and estimation, as well as for the evaluation and comparison of solution methods.

\subsection{Assessment and estimation}
\label{sec:assessment+estimation}  

Subproblems assessment and estimation have much in common.  In both cases data available is used in order to build an information landscape over a surface; in the latter case this landscape concerns the expected resource level at any given point in the surface, while in the former the landscape concerns the expected amount of information brought about if a given point is probed.

Methodologies for recommending points for data annotation with the objective to learn the target function are common in classification problems (see, \eg, \citet{ramirez2017active}); however, literature on this problem for regression is much scarcer.  One related problem is Bayesian optimization \citep{frazier2018tutorial, brochu2010tutorial}, where a point for evaluating a function is selected on the basis of the expectation of the improvement on the current best known solution, in the context of continuous optimization with expensive cost functions.  A measure of that expectation is conveyed by an \emph{acquisition function}, which is maximized at each iteration to determine where next to sample the objective function.

Both in our case and in Bayesian optimization the objective function is estimated by a surrogate function, derived from available observations of an (elsewhere unknown) ``true'' function and incorporating beliefs about its shape; but while in Bayesian optimization the aim is to observe points conducting to an optimum solution, in our case the objective is to have the surrogate function as close as possible to reality.

As in the case of Bayesian optimization, the surrogate function is determined by means of a \emph{Gaussian process regression} (see, \eg, \citet{Rasmussen2005}).  This is a generalization of the method known as \emph{kriging}, introduced in \citet{Krige1951} as a geostatistical procedure for generating an estimated surface from a scattered set of points with known values; the original application was on mine valuation.

Gaussian processes provide estimations for both the mean and the standard deviation; we use the latter as a measure of the attractiveness in the assessment phase, and the former as the value of the final estimation, after the data set is extended with points actually observed in the expedition.

Deciding the points to be probed is related to the problem of active learning; however, there are constraints on these points that must be taken into account, as probings must be done in an expedition whose length is limited.  This makes the connection between the assessment and estimation regression problems and the planning problem, \ie, the optimal schedule of a trip to visit them, which is discussed next.
}

\subsection{Planning}
\label{sec:planning}  
The optimization problem addressed in this work is a routing problem with similarities to the \emph{orienteering problem} (OP).  The orienteering problem was initially introduced by \citet{golden1981gtsp}, and its roots are in an outdoor sport with the same name, where there is a set of ``control points'', each with an associated score, in a given area.  Competitors use a compass and a map for assisting in a journey where they visit a subset of control points, starting and ending at given nodes, with the objective of maximizing their total score.  They must reach the end point within a predefined amount of time.

The input to the standard OP consists of a vertex- and edge-weighted graph $G = (V, E)$, a source and a target vertices $s, t \in V$, and a time limit $T$; $V$ is the set of vertices and $E$ is the set of edges.  The goal is to find an $s-t$ walk of total length at most $T$ so as to maximize the sum of weights at vertices visited through the walk.
It can be shown that the OP is NP-hard via a straightforward reduction from the traveling salesman problem.  It is also known to be APX-hard to approximate.  The literature describes an unweighted version (\ie, with a unit score at each vertex), for which a $(2+\epsilon)$ approximation is presented in \citet{Chekuri2012}; for the weighted version, the approximation ratio has a loss with factor $(1 + o(1))$.

An essential difference between the OP and our problem is that in the OP a finite set of vertices is given, from which the solution must be selected.  In our problem, only the surface where some locations may be chosen for sampling is given.  To the best of our knowledge, the closest related work can be found in \citep{Yu2016}. The authors propose a non-linear extension to the orienteering problem (OP), called the \emph{correlated orienteering problem} (COP). They use the COP to model the planning of informative tours for persistent monitoring, through a single or multiple robots, of a spatiotemporal field with time-invariant spatial correlations. The tours are constrained to have a fixed length time budget. Their approach is discrete, as they focus on a quadratic COP formulation that only looks at correlations between neighboring nodes in a network.

Another problem related to ours is \emph{tour recommendation for groups}, introduced in \citet{Anagnostopoulos2017}, where the authors deal with estimating the best tour that a group could perform together in a city, in such a way that the overall utility for whole group is maximized.  They use several measures to estimate this utility, such as the sum of the utilities of members in the group, or the utility of the least satisfied member.  In our case, we estimate this utility (the attractiveness of a point) using a Gaussian process regression, as described above.

\subsection{Related problems}
\label{sec:quality}
{ 
Methods for deciding data points whose unknown label should be determined are relatively common for classification tasks; see, \eg, \citet{pereira2017empirical} for a comparison.  However, to the best of our knowledge, no equivalent study has been done for regression problems.   A method for high-dimensional regression models, when the annotation of the samples is expensive, has been proposed in \citet{tsymbalov2018dropout}.  There, the authors present a fast active learning algorithm for regression, tailored for neural network models; the stochastic output of the neural network model, performed using different dropout masks at the prediction stage, is used to rank unlabeled samples and to choose among them the ones with the highest uncertainty.  Even though the setting is somehow similar to ours, the pool of unlabeled points is considered a finite set; these points are ranked at each iteration according to the respective value of the acquisition function, and those with the best value are chosen.  Hence, data sets from standard machine learning can be used in their context.
In our context, any point in the surface could potentially be selected for annotation; hence, common data sets for regression problems, where labels only exist for a discrete subset of points in the domain, are not applicable.  Besides, in our context data available is supposed to be scarce; otherwise regression without acquisition of new points would probably be satisfactory since the beginning, and hence the interest of an expedition for collecting new data would be limited.

In the context of exploration problems, methods for using novelty in robotic exploration have been studied in \citet{thompson2007predictive}, where an algorithm based on density estimation and adaptive threshold detection is proposed to characterize novelty, taking into account a prior model of novel data.  It has been used with images from a particular mission, where the algorithm identifies interesting cases among a variety of images.
On a different perspective, a discussion on the usage of machine learning techniques in exploration of unknown places---rovers on Mars surface---has been made in \citet{castano2003machine}.  The usage of quantitative measurements that can match the algorithms to the priorities of experts is proposed. Applications include the situation where image-based novelty can be used \eg~for prioritizing images to be transmitted, or to provide the rover with the capability of changing its course or stopping when a key factor is detected, in order to improve the ``science return''  of the mission.
The main difference between these works and our setting is that prior information about the surface being explored is not taken into account for scheduling the expedition.  Besides, in these previous works it is assumed to have a continuous flow of information, while in our case probing must be made on a discrete set of points; in other words, if we consider an underlying graph, these works collect information on arcs, while in the sea exploration problem information is collected on vertices.  The position of these vertices and the corresponding order of visit are the variables at stake.

Another related problem is that of searching environments in rescue operations.  This problem is dealt with in~\citet{basilico2011exploration}, which presents a multi-criteria method for searching an environment without any \emph{a priori} information about structure and target locations.  There, the authors combine a set of criteria (amount of anticipated free area, probability of a robot being able to send back information, distance between the current position and the place to explore) in order to produce an assessment of the global utility for a candidate point for exploration;  the point which maximizes this utility is visited next (see also \citet{li2015study} for a survey on related problems).  Contrasting to this problem, in our setting the aim is to use previously available information and improve it in an optimal way.
}

\section{Method}
\label{sec:method}

{ 
  A general overview of the main algorithm, showing the information flow among the main steps---assessment, planning and estimation---is presented in Figure~\ref{fig:algorithm}.  This section provides a detailed description of each of these steps.  All the components of this algorithm are deterministic; potentially, some procedures called by our method could fail to consistently find the same solution, but we did not observe such behavior in the experiments.
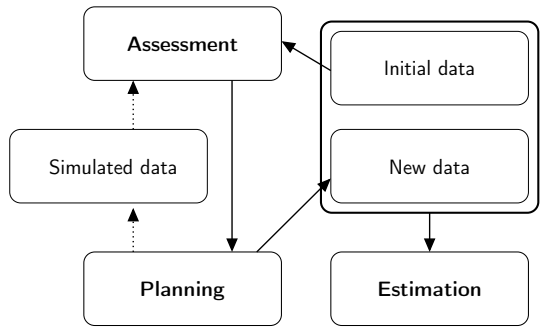
\begin{figure}[!htbp]
  \begin{center}
  \scalebox{.65}{\begin{tikzpicture}[y=-1cm]

\draw[rounded corners=6.3bp,semithick,black] (11,6.5) rectangle (7,5);
\draw[rounded corners=6.3bp,black] (11,4) rectangle (7,2.5);
\draw[rounded corners=6.3bp,black] (11,2) rectangle (7,0.5);
\draw[rounded corners=6.3bp,semithick,black] (6,1.5) rectangle (2,0);
\draw[rounded corners=6.3bp,very thick,black] (11.2,4.2) rectangle (6.8,0.3);
\draw[rounded corners=6.3bp,black] (4.5,4) rectangle (0.5,2.5); 
\draw[rounded corners=6.3bp,semithick,black] (2,5) rectangle (6,6.5); 

\draw[thick,arrows=-triangle 45,black] (5,1.5) -- (5,5); 
\draw[thick,arrows=-triangle 45,black] (5.5,5) -- (7,3.5);
\draw[thick,arrows=-triangle 45,black] (9,4.2) -- (9,5);
\draw[thick,arrows=-triangle 45,black] (7,1.3) -- (6,0.7);
\draw[thick,dotted,arrows=-triangle 45,black] (3,5) -- node[above,rotate=90] {} (3,4.1);
\draw[thick,dotted,arrows=-triangle 45,black] (3,2.5) -- node[above,rotate=90] {} (3,1.5);

\path (4,0.9) node[text=black,anchor=base] {\large{}\sffamily \textbf{Assessment}};
\path (9,3.4) node[text=black,anchor=base] {\large{}\sffamily New data};
\path (9,1.4) node[text=black,anchor=base] {\large{}\sffamily Initial data};
\path (2.6,3.4) node[text=black,anchor=base] {\large{}\sffamily Simulated data};
\path (4.0,5.9) node[text=black,anchor=base] {\large{}\sffamily \textbf{Planning}};
\path (9,5.9) node[text=black,anchor=base] {\large{}\sffamily \textbf{Estimation}};

\end{tikzpicture}%
} 
  \end{center}
\caption{Information flow and interactions between different parts of the algorithm.}
\label{fig:algorithm}
\end{figure}
}

\subsection{Assessment and estimation.} The first and the third subproblems, assessment and estimation, are strongly related, in the sense that the aim of the assessment phase is to have a measure of the interest of having empirical information on new points in~$S$ for improving the estimation.  Assessment evaluates how much a given point, if probed, is expected to improve the quality of the estimation done at the third subproblem.

The estimation phase is a regression problem: given the known resource levels at the $N$ previously observed points and at $n$ points to be observed in the expedition, what is the best estimate for the resource level at a new point in~$S$?  A first step for answering this question is to make an assumption on the nature of the underlying function $v(x,y)$.  Our assumption is that it can be conveniently estimated by a Gaussian process.  In this approach, the model attempts to describe the conditional distribution $p(z|(x,y))$ based on a set of empirical observations of $z$ on input $(x,y)$, conveyed as a set of triplets $D = \{(\dz_i,\dx_i,\dy_i)\}_{i=1}^{m}$, where $m$ is the number of samples (in our case, $N$ before the trip and $N+n$ after the trip).   This conditional distribution describes the dependency of the observable $z$ on the input $(x,y) \in S$, assuming that this relation can be decomposed into a systematic and a random component.  The systematic dependency is given by a latent function $w : S \to \mathbb{R}$, which is to be identified based on data $D$.  Hence, the prior is on function values associated with the set of inputs, whose joint distribution is assumed to be multivariate normal.

We use the posteriors inferred through the Gaussian process model in two ways.  In the assessment phase, we use the standard deviation of the model at each point $(x,y) \in S$ directly as an indicator of attractiveness for probing at that point.  Later, in the estimation phase, the Gaussian process (now with an enlarged data set) is used as a regression for the resource level at any point in~$S$.

\subsection{Planning.} The second subproblem is the selection of points in $S$ for probing, so as to allow a subsequent estimation as accurate as possible.  Points are to be probed in a trip whose maximum duration is known beforehand.  We are thus in the presence of an orienteering problem.  A standard orienteering problem consists of the following: given a graph with edge lengths and a prize that may be collected at each vertex, determine a path of length at most $T$, starting and ending at given vertices, that maximizes the total prize value of the vertices visited.  The problem here is rather particular for several reasons.  The first reason is that besides edge lengths (in our case, edge traversal duration), we have to take into account the time spent in probing at each vertex (which is a parameter of our problem).
The second reason is that the graph may consist of any discrete subset of points $V \subset S$, as long as the duration of the tour --- the time spent on probing and on traveling from a point to the next --- is not larger than the upper bound~$T$.  An additional difficulty is related to the correlation between the prizes obtained in visited vertices; indeed, as the ``prize'' is a measure of the improvement on information obtained by probing, after probing at a given location, probing other locations in this neighborhood is expected to provide less information than distant points (other factors being equal).

\subsection{Tackling the problem}
An instance of this problem must specify the area $S$ being studied, an upper bound~$T$ for the trip duration (including traveling and probing), the duration $t$ required for each probing (here considered independent of the location), and the traveling speed $s$, which allows computing the traveling time between two given points as $d/s$, where $d$ is the Euclidean distance between those points.  Without loss of generality, we are assuming that the initial and end points of the trip to be planned are the same.  Among the instance's data, it must also be provided the previously known data $D = \{(\dz_i,\dx_i,\dy_i)\}_{i=1}^{N}$, corresponding to a set of points $(\dx_i,\dy_i) \in S$ and the corresponding resource level $\dz_i$, for $i=1,\ldots,N$.

In order to evaluate an algorithm for this problem, another set of $K$ points $E = \{(\sx_k,\sy_k)\}_{k=1}^{K}$ at which the level of information predicted by the model is requested, should also be specified.  For these points, the true value of the resource level $v(\sx_k,\sy_k)$ must be known at the end (for computing the error with respect to its estimated value); notice, however, that these values cannot be used by the algorithm.

{ 
The main course of action, making use of a set of auxiliary functions, is provided in Algorithm~\ref{alg:main}.   A general view of its steps is the following.
In line 1 
the relevant data necessary for the algorithm to generate a solution is input: previously available data $D$, time budget $T$, probing time $t$, ship's speed $s$ and surface under study~$S$.  In order to highlight that only previously available data $D$ can be used by the method, the ``true function'' $v$ and the points at which the error will be evaluated $E$ are only input after computing the solution, in lines 3 and 6, respectively. 
Then, the solution is computed in line 2 
by means of Algorithm~\ref{alg:orienteering} described below.  This algorithm returns the list of points to probe, in the order of visit, so that the time limit $T$ is not exceeded.  The true function is then evaluated at these points (line 4), 
and the updated set of points $D$ is then used to train a Gaussian process (line 5) 
whose mean evaluation at each point $(x,y) \in E$ is compared to function $v$ (line 7), 
where $E$ is the set of points used for error evaluation.
}
\begin{algorithm}[!htbp]
  \begin{footnotesize}
    \DontPrintSemicolon
    \SetKwFunction{Orienteering}{Orienteering}
    \SetKwFunction{GPstdev}{GPstdev}
    \SetKwProg{myproc}{procedure}{}{}
    read instance's data $D, T, t, s, S$\;
    $a \eq \Orienteering(D, T, t, s, S)$ \;
    input instance's ``true function'' evaluator $v$\;  
    update $D$ with probings on all points $a$\;
    $w \eq $ GP regression trained with data $D$\;
    input points for error evaluation $E$\;  
    output $\Delta \eq \sum_{(x,y) \in E} {|v(x,y) - w(x,y)|}$\;
  \end{footnotesize}
  \caption{Main procedure.}
  \label{alg:main}
\end{algorithm}

{ 
  The main heuristic concept behind the algorithm is that by adding to the data set $D$ points where the variance (and hence the uncertainty) is large, and using them for determining the function that approximates reality (or, in the benchmark tests, the function that approximates~$v$), its outcome will be best improved.  Hence, an estimation of the variance (or, equivalently, of the standard deviation) is necessary; this is obtained in Algorithm~\ref{alg:attract} by fitting a Gaussian process to the (growing) set of points $D$ (line 2) 
  and computing its standard deviation at a set of points $G$ covering the relevant surface~$S$.  In this auxiliary algorithm, these points are then sorted and returned (lines 4 and 5); 
  the best of them (\ie, the last in their ordered set) will be selected.

  One difficulty mentioned earlier in this section concerns the correlation between variances observed at points in the surface.  This is noticeable in the ordered set returned by Algorithm~\ref{alg:attract}: its tail is likely to be composed of points which have a similar high variance, but which lie next to each other in the surface~$S$.  Hence, upon selection of one (the best) of them for including in the solution, the Gaussian process used to compute attractiveness is unusable; a new one must be fitted to the data, including that last point added.  These concerns have been addressed in Algorithm~\ref{alg:orienteering}, described next.
}
\begin{algorithm}[!htbp]
  \begin{footnotesize}
    \DontPrintSemicolon
    \SetKwFunction{length}{length}
    \KwData{Data: $D = \{(\dz_i,\dx_i,\dy_i)\}_{i=1}^{N}, T, t, s, S$}
    \KwResult{List $a$ of points to visit for probing}
    \SetKwProg{myproc}{procedure}{}{}
    \SetKw{Break}{break}
    \SetKw{Continue}{continue}
    \SetKw{True}{True}
    \SetKw{pop}{pop}
    \SetKw{append}{append}
    \myproc{\Orienteering{$D, T, t, s, S$}}{
      $a \eq []$ \tcp*[f]{list of points to probe}\;
      \While{\True}{
        $V \eq \GPstdev(D,S)$ \tcp*[f]{points on $G$ sorted by attractiveness} \;
        $(\sigma,x,y) \eq V.\pop()$  \tcp*[f]{remove and return last element of $V$} \;
        $r \eq $ TSP solution visiting $(x,y)$ and all the positions in $a$\;
        \If(\tcp*[f]{including time for probing}){\length(r) < T}
        {
          $a \eq r$
        }
        \Else{\Break}
        $R \eq $ GP regression trained with data $D$\;
        $z \eq R(x,y)$\;
        $D.\append((z,x,y))$\;
      }
      \Return $a$
    }
  \end{footnotesize}
  \caption{Orienteering.}
  \label{alg:orienteering}
\end{algorithm}

Algorithm~\ref{alg:orienteering} uses the assessment of the attractiveness on a grid of points in the surface~$S$ to determine a trip, \ie, a list of points to visit and probe.  That list is constructed in a greedy way, by determining which is currently the most attractive point---by calling Algorithm~\ref{alg:attract} in line 4---and 
attempting to add it to the trip, in lines 6 to 10. 
This is done by checking if a traveling salesman tour including it and the previous points can still be completed within the time limit (we use an implementation of the algorithm described in \citet{Lin1973} for quickly finding a tour; if its length is feasible, the solution is immediately returned, otherwise the exact model available in \citet{pedroso2012gurobi-book} is used to find the optimal solution with a general-purpose mixed integer programming solver).
{ 
  Notice that in the length of the tour one must include the travel time between successive points and the time for probing at each point.
}
After a new point $(x,y)$ is added to the tour, it is conjectured (for the purposes of the algorithm) that a simulation using the latest available Gaussian process provides the ``true'' evaluation $z = v(x,y)$; based on the new speculative datum $(z,x,y)$ the attractiveness allover $S$ will be recomputed in next call to Algorithm~\ref{alg:attract}.
\begin{algorithm}[!htbp]
  \begin{footnotesize}
    \DontPrintSemicolon
    \SetKwFunction{sort}{sort}
    \KwData{Data: $D = \{(\dz_i,\dx_i,\dy_i)\}_{i=1}^{m}, S$}
    \KwResult{Assessment of standard deviation on grid $G$, $\{(\sigma_i,x_i,y_i)\}_{i\in G}$}
    \SetKwProg{myproc}{procedure}{}{}
    \SetKw{Break}{break}
    \SetKw{Continue}{continue}
    \myproc{\GPstdev{$D, S$}}{
      $G \eq \{(x_0+\delta k, y_0+\delta \ell), k=0, \ldots, K, \ell=0, \ldots, L\}$ \tcp*[f]{~\hspace{-1mm}grid on $S\!\!=\!\![0,1]^2\!\!: \delta\!\!=\!\!\frac{1}{K}\!\!=\!\!\frac{1}{L}$} \;  
      $R \eq $ GP regression trained with data $D$\;
      $s \eq $ standard deviation function, as evaluated by $R$\;
      $V \eq \{(s(x_i,y_i), x_i, y_i)\}_{i \in G}$ \;
      \sort(V)\;
      \Return V\;
    }
  \end{footnotesize}
  \caption{``Attractiveness'' of points for probing.}
  \label{alg:attract}
\end{algorithm}

\section{Computational results}
\label{sec:results}

{ 
Ideally, the quality of a method for the sea exploration problem would be determined by its prediction error, \ie, by the difference between the true seafloor contents $v(x,y)$ and the contents predicted by the model $w(x,y)$, at any arbitrary point $(x,y)$.  If we denote this error by $e(x,y) = v(x,y) - w(x,y)$, then the ideal measure of the method's error $R$ is its integral over the relevant surface~$S$:
$$R = \iint_S |e(x,y)| dx dy$$
In our problem, as in many practical situations, we do not have access to the ground truth allover $S$; hence, $R$ must be somehow approximated.  We devised a benchmarking process based on artificial ``true'' functions which allows us to compare different methods; we compare simple grid search to the algorithm proposed in the previous section.
}

\subsection{Benchmark instances used}

In order to assess the quality of the method proposed for solving this problem, we have devised a set of artificial benchmark instances (see Appendix~\ref{sec:instances} for details on their characteristics).

{ 
Notice that our algorithm decides the position on the relevant surface of a set of points for probing; hence, testing the quality of an algorithm is only possible if the ``ground truth'' is known on a continuum.  As a replacement for this ``ground truth'', we devised a test set consisting of 10 functions of varying complexity.  In each benchmark, an initial small set of points on the surface and the corresponding value of the function are given; these are used in the assessment step.  Further points, chosen by the algorithm at the planning step, will be probed at the end of this step---\ie, their function value will be calculated.  These two sets will be used at the estimation step for building the function that will approximate the ground truth.  Finally, differences between these two functions, evaluated at another set of points over the surface, will be summed and used to determine the error of the algorithm.  With this setting we can establish the quality of an algorithm, and use it to compare different algorithms.
}

A solution to our problem consists of two parts: a sequence of locations $a = [(\tx_1,\ty_1), \ldots, (\tx_n,\ty_n)]$ for points where to collect samples (aiming at having maximum information collected), such that the total probing and travel time does not exceed~$T$; and an estimation $\sz_k$ of the level of information predicted by the model on each of the $K$ requested points $E = \{(\sx_k,\sy_k)\}_{k=1}^{K}$.

The evaluation of a method for solving this problem is firstly based on feasibility: the orienteering tour is checked by verifying that the duration of tour $[(\tx_1,\ty_1), \ldots, (\tx_n,\ty_n), (\tx_1,\ty_1)]$ does not exceed~$T$.  Then, the final solution quality is measured by $\sum_{k=1}^{K} {|v(x_k,y_k) - w(x_k,y_k)|}$, as an approximation of $\iint |v(x,y) - w(x,y)| dx dy$.

\subsection{Results obtained}

{ 
The computer platform used was an Intel Core i7 quad-core CPU, running at 3.4  GHz in a Mac OS X version 10.10.2, with 24 GB of RAM.  All programs were implemented in Python (version 2.7.9), using Scikit-learn toolkit version 0.18.2 \citep{scikit-learn} and Gurobi optimizer version 7.0 \citep{gurobi}.
}

Tables~\ref{tab:res16a} to~\ref{tab:res100b} report the results of the computational experiment executed.
Instances for the following cases have been solved: 16 previously known points on a regular grid (Table~\ref{tab:res16a}) and on random positions (Table~\ref{tab:res16b}); 
instances with 49 points on a regular grid (Table~\ref{tab:res49a}) and on random positions (Table~\ref{tab:res49b}); and 
instances with 100 points on a regular grid (Table~\ref{tab:res100a}) and on random positions (Table~\ref{tab:res100b}).
Instances 1 to 5 have an increasing number of local maxima, but are relatively smooth; instances 6 to 10 are less so, with some of them having rather narrow local maxima.
Parameters used in the Gaussian process were roughly tuned using benchmark instances \texttt{f1} to \texttt{f5}; hence, these instances can be seen as the training set, and instances \texttt{f6} to \texttt{f10} as the test set.

In order to have an assessment of the quality of Algorithm~\ref{alg:orienteering}, we compare it to a simple grid search, dividing the time available for exploration into probes on a regular grid, simply avoiding new observations on points for which there was already information available (notice that when the previously available information is very scarce, searching on a grid is a sensible strategy).  Both methods use the same Gaussian process for regression; hence, their initial solution is the same.

As expected, increasing the number of initial points with information available increases the quality of the initial estimation with a Gaussian process; this can be observed in the first column of tables~\ref{tab:res16a} to~\ref{tab:res100b}.  Having those points disposed in a grid is usually preferable, but this is not a general pattern.  The estimation is much improved when new data becomes available (an exception is observed on instance \texttt{f7} with 100 initial points randomly spread, where adding more data on a regular grid strangely decreased the quality of the final estimation; this is likely to be a limitation of the optimizer used for training the Gaussian processes).

The main results 
%
{ 
  are presented in tables~\ref{tab:res16a} to~\ref{tab:res100b} and summarized in Table~\ref{tab:results}.
  A winning case reported in Table~\ref{tab:results} corresponds to the situation where an algorithm obtained a lower error than the other (according to the metric defined in the beginning of this section).  Hence, larger values in this table correspond to a better algorithm.
}
%
We can observe that the orienteering method proposed in Algorithm~\ref{alg:orienteering} is generally superior to grid search.
This superiority increases with the quantity of initial points available, indicating that, \eg, 16 previous points do not provide enough information for preparing a probing plan based on those data.

\begin{table}[!htbp]
  \centering
  \caption{Summary of results: winning situations for grid search and for our algorithm.}
  \begin{tabular}{lccc}
    Benchmarks & Grid search & Orienteering \\\hline
    16-grid    &      4      &    6      \\
    49-grid    &      2      &    8      \\
    100-grid   &      2      &    8      \\
    16-random  &      6      &    4      \\
    49-random  &      5      &    5      \\
    100-random &      1      &    9      \\
  \end{tabular}
  \label{tab:results}
\end{table}

\begin{table}[!htbp]
  \centering
  \caption{Results for instances with 16 initial points on a regular grid.}
\sisetup{
round-mode = places,
round-precision = 1 }%
\begin{tabular}{r|SSSS}
  \toprule
  {Name} & {Initial} & {Grid} & {Orienteering} \\
  \midrule
  f1  & 37179.778811730626 &    911.7041527629668 & 622.5733146415533   \\ 
  f2  & 40696.71629611851  &   7509.139481896236  & 6094.044365285448   \\ 
  f3  & 52257.19063109544  &   6441.789703795481  & 7935.324447441893   \\ 
  f4  & 60996.87798721852  &   9768.986380976852  & 14253.397340910882  \\ 
  f5  & 61339.81507957516  &   9910.764540742326  & 9415.666057980878   \\ 
  f6  & 56.15433842301819  &      1.3856423092470 & 1.7802725274184277  \\ 
  f7  & 730.0779372614868  &   686.2554885474509  & 675.8285069442941   \\ 
  f8  & 34764.77470303922  &  1415.7205093903674  & 975.096960919084    \\ 
  f9  & 55643.53539494152  &  2116.000166529825   & 2007.9100295870455  \\ 
  f10 & 57428.169869564044 &  4547.918542351984   & 6306.045952962949   \\
  \bottomrule
\end{tabular}
  \label{tab:res16a}
\end{table}

\begin{table}[!htbp]
  \centering
  \caption{Results for instances with 16 initial points on random positions.}
\sisetup{
round-mode = places,
round-precision = 1 }%
\begin{tabular}{r|SSSS}
  \toprule
  {Name} & {Initial} & {Grid} & {Orienteering} \\
  \midrule
  f1  & 45327.99223 & 981.0667238753 & 393.0387719508  \\ 
  f2  & 34509.91816 & 6874.414139594 & 8629.612791418  \\ 
  f3  & 48476.62221 & 7282.731103267 & 11217.08391896  \\ 
  f4  & 40291.26654 & 11324.12807287 & 13683.13276886  \\ 
  f5  & 67108.55182 & 8293.709629573 & 13505.75216142  \\ 
  f6  & 51.22287120 & 1.526349702613 & 1.855124146473  \\ 
  f7  & 710.9913316 & 686.6961762270 & 680.3497195064  \\ 
  f8  & 44127.14824 & 1414.289432736 & 1138.942175462  \\ 
  f9  & 62934.96119 & 2048.881168471 & 5661.822273599  \\ 
  f10 & 39025.07259 & 5966.048960339 & 5711.869140318  \\
  \bottomrule
\end{tabular}
  \label{tab:res16b}
\end{table}

\begin{table}[!htbp]
  \centering
  \caption{Results for instances with 49 initial points on a regular grid.}
\sisetup{
round-mode = places,
round-precision = 1 }%
\begin{tabular}{r|SSSS}
  \toprule
  {Name} & {Initial} & {Grid} & {Orienteering} \\
  \midrule
  f1  & 12108.97137666 & 181.731601068111 & 908.18259137810  \\ 
  f2  & 19311.70659492 & 4443.06488192847 & 4312.2358937250  \\ 
  f3  & 24787.39796680 & 5159.35433327471 & 4698.6601148211  \\ 
  f4  & 24215.69612876 & 5197.24636896981 & 4915.5596035177  \\ 
  f5  & 24157.86703339 & 5358.61322013083 & 5047.5866646386  \\ 
  f6  & 5.086410579774 & 1.28152906355877 & 1.5002020919382  \\ 
  f7  & 2442.656906440 & 1186.07508274749 & 1030.0429129309  \\ 
  f8  & 14703.08318244 & 2304.45314508131 & 1414.8509135940  \\ 
  f9  & 23072.05523854 & 2811.45162635772 & 1993.6269417231  \\ 
  f10 & 25322.55762799 & 5009.43979135893 & 4032.6911555676  \\
  \bottomrule
\end{tabular}
  \label{tab:res49a}
\end{table}

\begin{table}[!htbp]
  \centering
  \caption{Results for instances with 49 initial points on random positions.}
\sisetup{
round-mode = places,
round-precision = 1 }%
\begin{tabular}{r|SSSS}
  \toprule
  {Name} & {Initial} & {Grid} & {Orienteering} \\
  \midrule
  f1  & 13997.25353754 & 480.36351574704 & 733.2685926695  \\ 
  f2  & 17605.45853853 & 3267.8122742294 & 6330.861960372  \\ 
  f3  & 28619.33116749 & 4357.5807145301 & 3726.929905739  \\ 
  f4  & 35975.46408136 & 7740.8285866607 & 6191.163191926  \\ 
  f5  & 45076.35307099 & 9694.1523334896 & 6937.110554097  \\ 
  f6  & 10.44123444584 & 1.2535376537810 & 1.475482027571  \\ 
  f7  & 678.3053485673 & 685.52857459377 & 675.9856747658  \\ 
  f8  & 23555.57918052 & 826.83341088183 & 752.7486088134  \\ 
  f9  & 22699.31402901 & 1893.8774797966 & 2837.814343250  \\ 
  f10 & 18552.67478668 & 3835.0922993059 & 4540.825450144  \\
  \bottomrule
\end{tabular}
  \label{tab:res49b}
\end{table}

\begin{table}[!htbp]
  \centering
  \caption{Results for instances with 100 initial points on a regular grid.}
\sisetup{
round-mode = places,
round-precision = 1 }%
\begin{tabular}{r|SSSS}
  \toprule
  {Name} & {Initial} & {Grid} & {Orienteering} \\
  \midrule
  f1  & 1731.7896272028 & 44.05442596446 & 13.981762013760  \\ 
  f2  & 6426.9197973531 & 3221.388218697 & 2709.2257612550  \\ 
  f3  & 8809.9251941359 & 4060.648972336 & 2801.2783635260  \\ 
  f4  & 11401.920965345 & 5408.026716541 & 4552.7476132560  \\ 
  f5  & 11740.116089340 & 5584.462915850 & 5103.4756775125  \\ 
  f6  & 2.1472886022932 & 1.229213042754 & 1.2623032999025  \\ 
  f7  & 4889.2470181203 & 3015.201113918 & 3714.0467289974  \\ 
  f8  & 10252.211519105 & 5633.021459386 & 4140.5621072140  \\ 
  f9  & 13115.303241305 & 6356.344860132 & 4435.7027588911  \\ 
  f10 & 14806.818020229 & 7249.516893536 & 6028.6725916451  \\
  \bottomrule
\end{tabular}
  \label{tab:res100a}
\end{table}

\begin{table}[!htbp]
  \centering
  \caption{Results for instances with 100 initial points on random positions.}
\sisetup{
round-mode = places,
round-precision = 1 }%
\begin{tabular}{r|SSSS}
  \toprule
  {Name} & {Initial} & {Grid} & {Orienteering} \\
  \midrule
  f1  & 770.81741855039 & 15.0984707795 & 12.4006697730  \\ 
  f2  & 16669.033357815 & 5057.86816844 & 2904.09709395  \\ 
  f3  & 16099.794024674 & 5152.52474174 & 4009.43488407  \\ 
  f4  & 17974.570559110 & 6475.31068938 & 3115.24569124  \\ 
  f5  & 25562.243575807 & 4947.56294429 & 4131.39991938  \\ 
  f6  & 1.5818950001535 & 1.08188582527 & 1.54268838260  \\ 
  f7  & 675.68542582178 & 682.798544398 & 675.419307010  \\ 
  f8  & 883.83751620512 & 6666.55104402 & 2555.98483683  \\ 
  f9  & 3736.2339833565 & 2079.54693306 & 837.336643633  \\ 
  f10 & 9484.1353914149 & 3338.82099048 & 3018.13328387  \\
  \bottomrule
\end{tabular}
  \label{tab:res100b}
\end{table}

\begin{figure}[h!tbp]
  \centering
  \includegraphics[width=.32\textwidth,trim=100 30 50 50,clip=True]{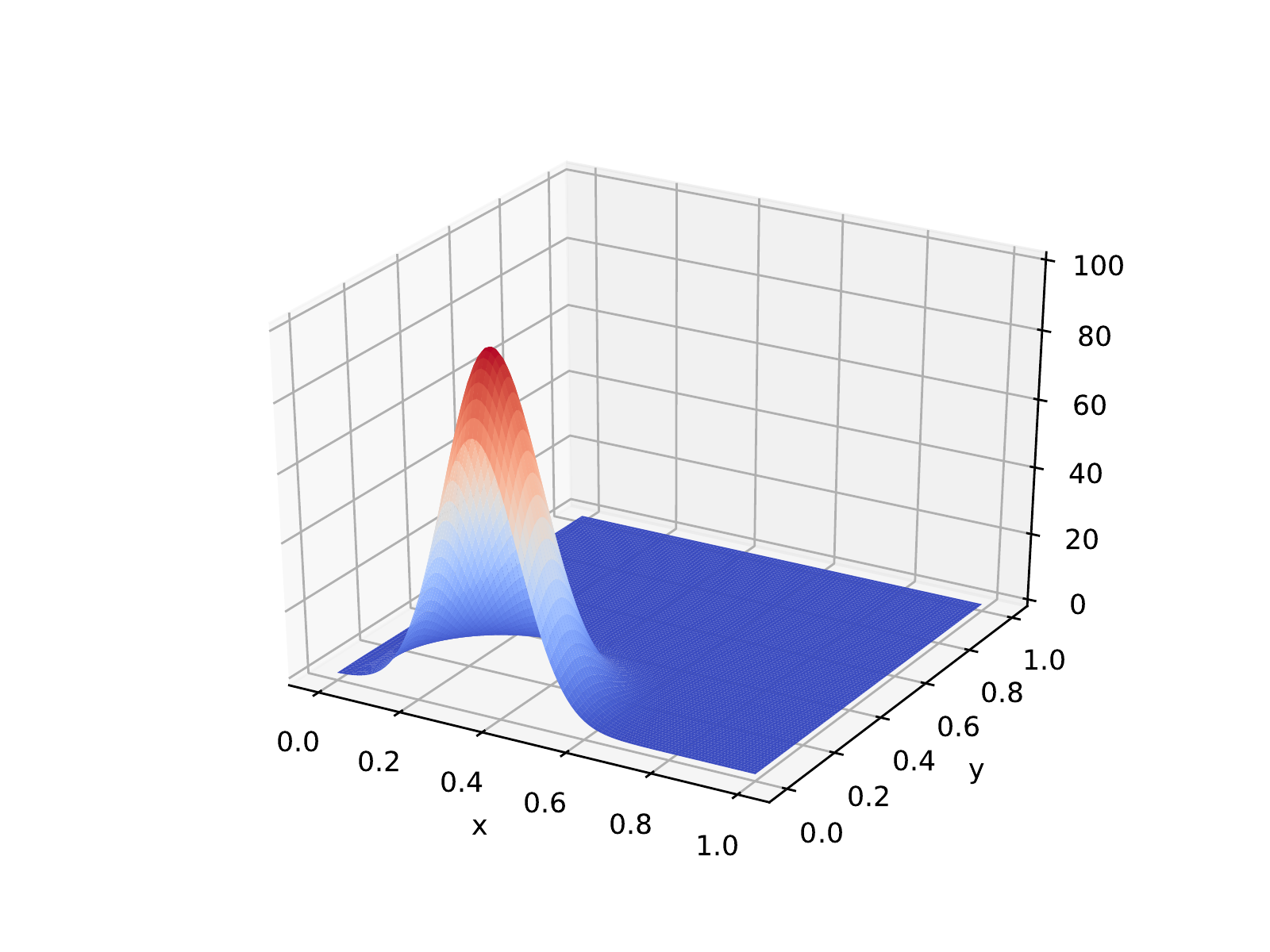}
  \includegraphics[width=.32\textwidth,trim=100 30 50 50,clip=True]{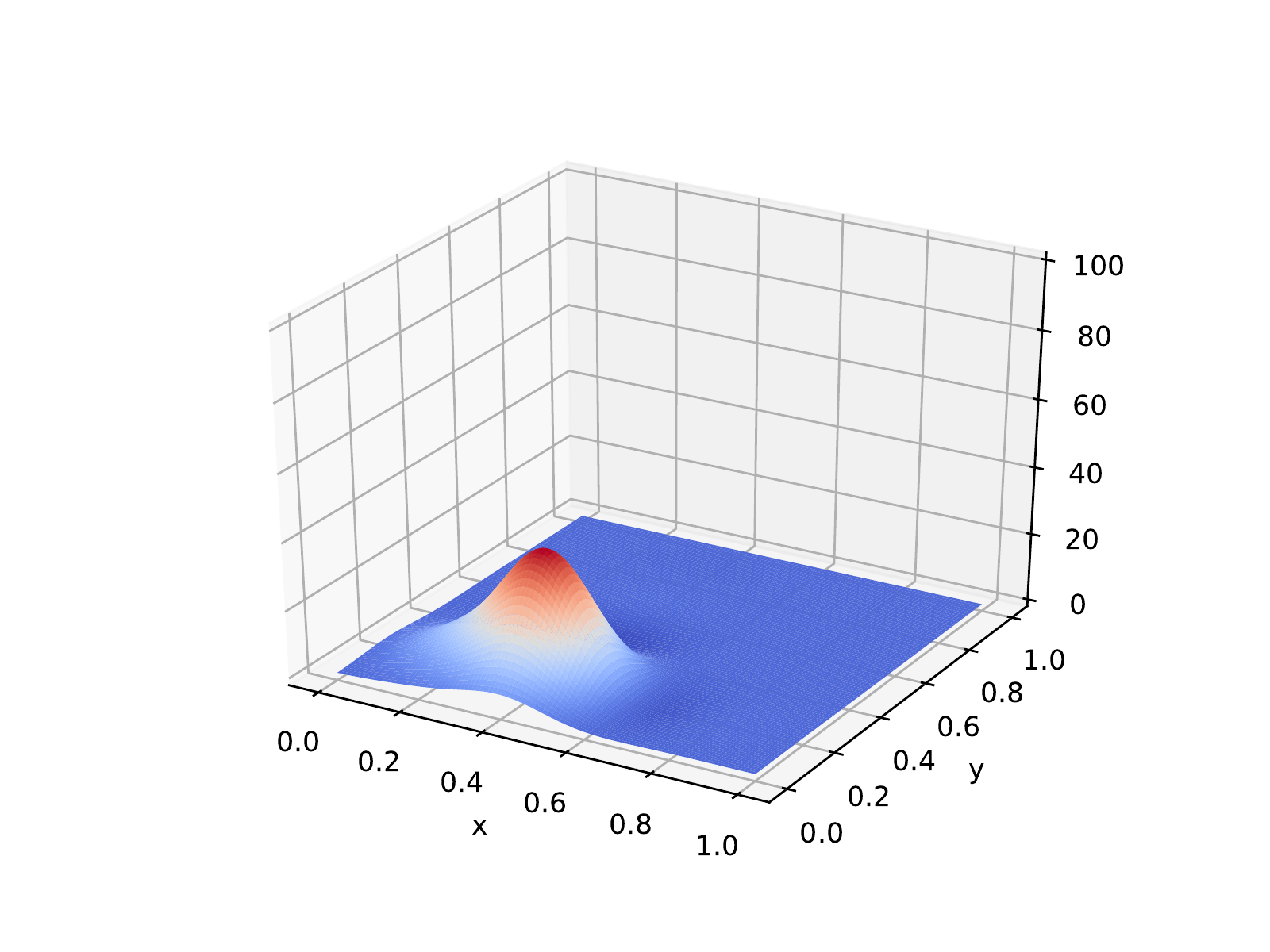}
  \includegraphics[width=.32\textwidth,trim=100 30 50 50,clip=True]{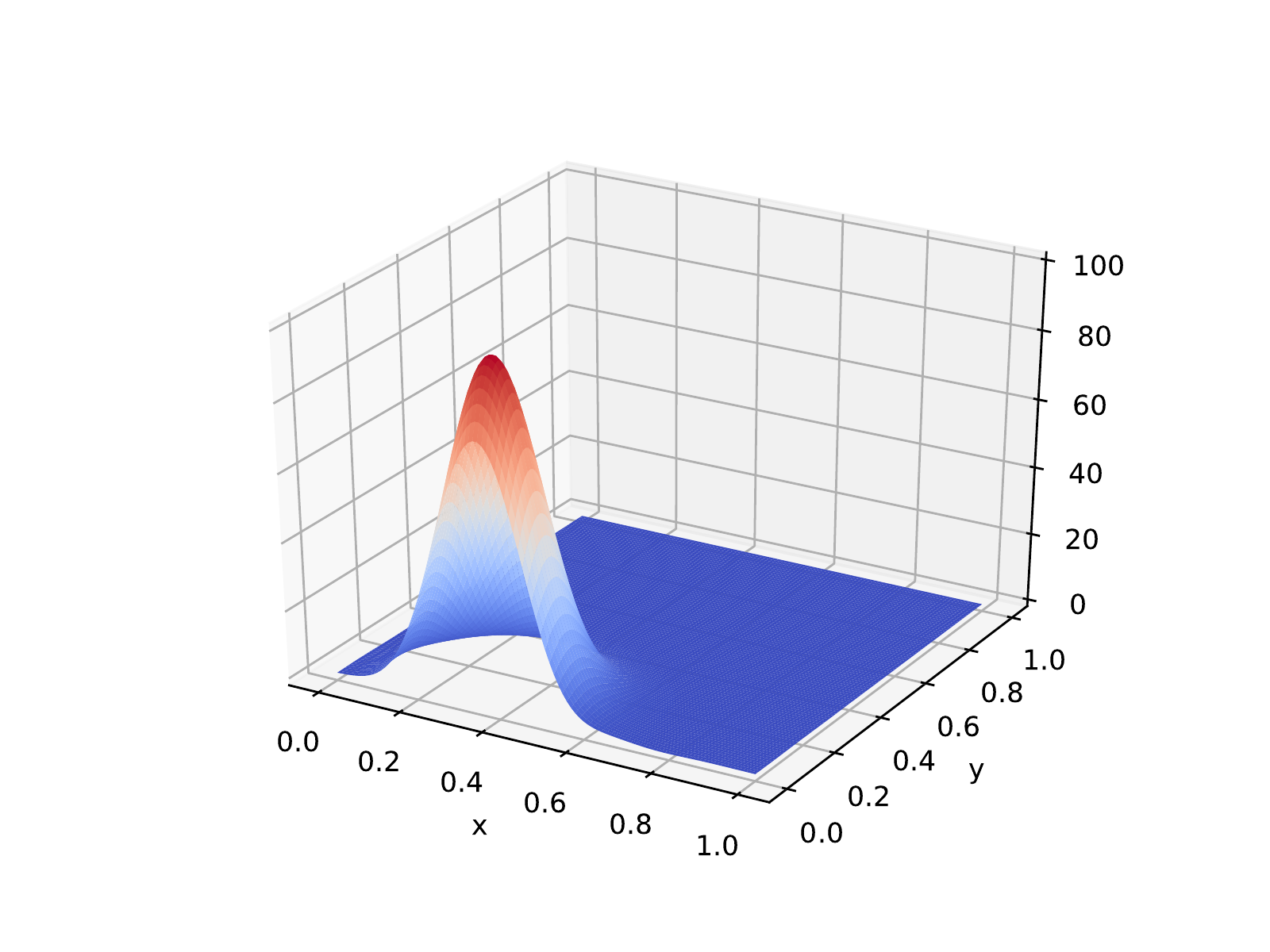}
  \caption{``True'' function used in instance \texttt{f1} (left), initial estimation (center) and estimation after probing (right).}
  \label{fig:f1}
\end{figure}

\begin{figure}[h!tbp]
  \centering
  \includegraphics[width=.32\textwidth,trim=100 30 50 50,clip=True]{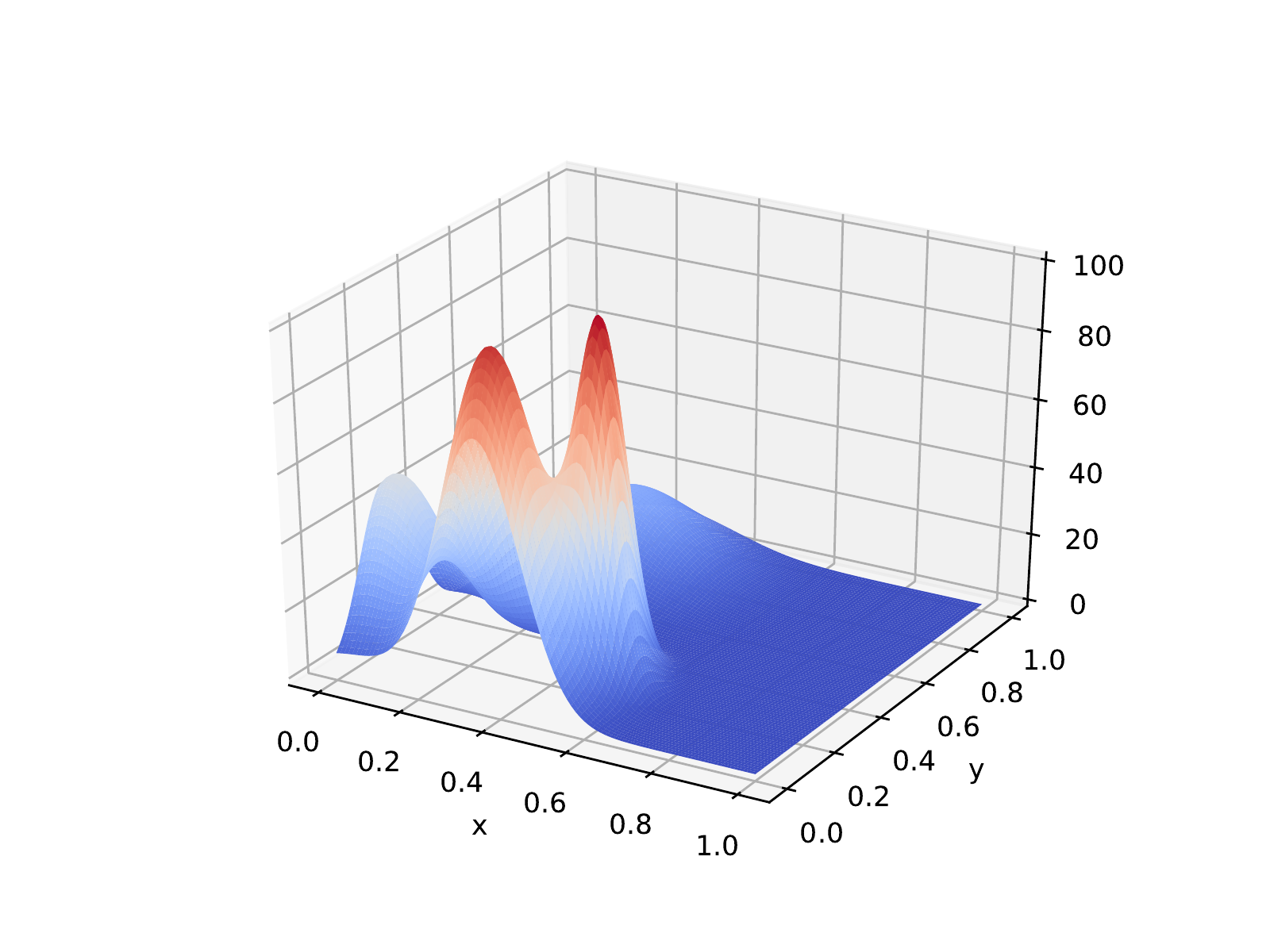}
  \includegraphics[width=.32\textwidth,trim=100 30 50 50,clip=True]{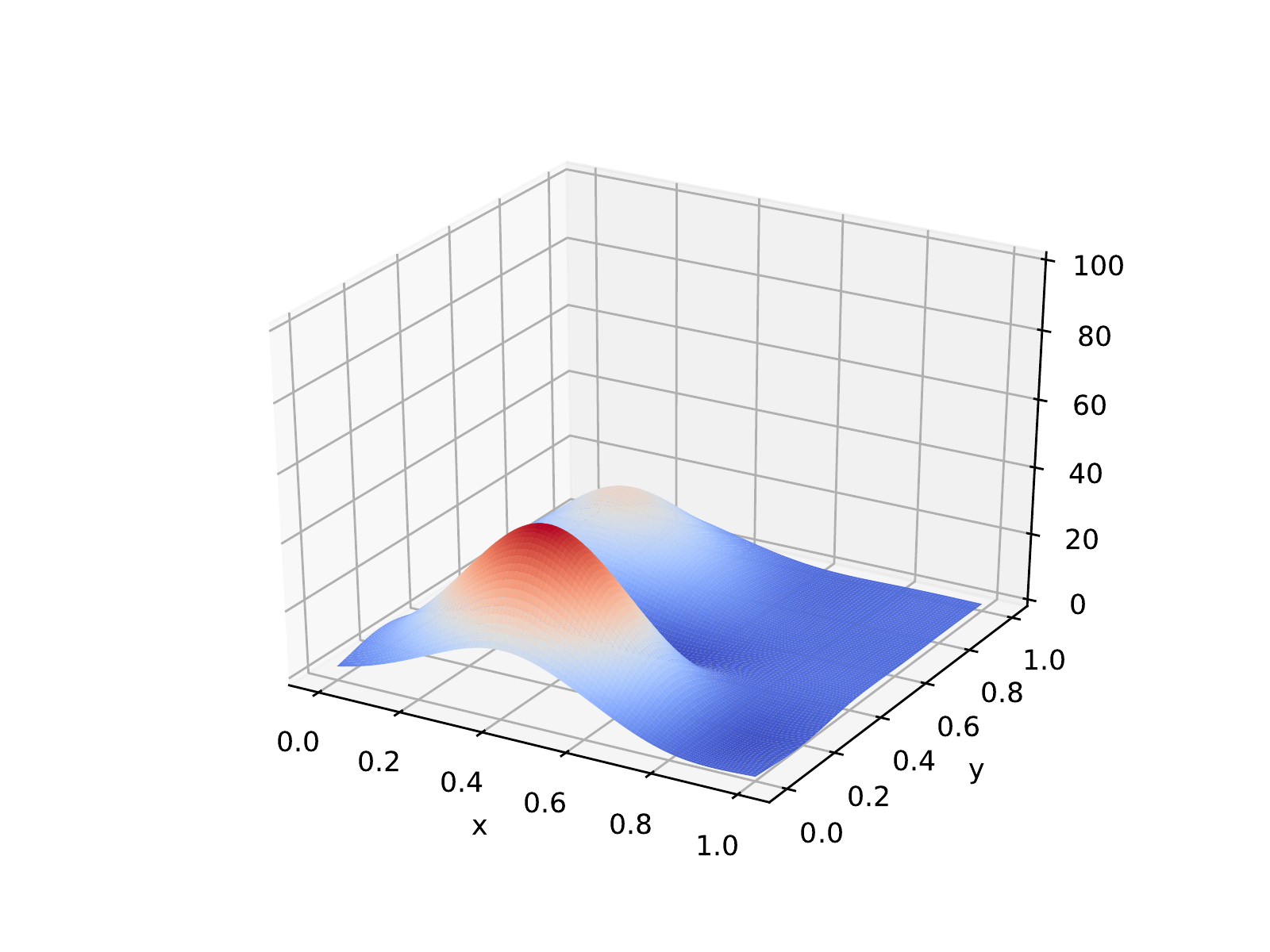}
  \includegraphics[width=.32\textwidth,trim=100 30 50 50,clip=True]{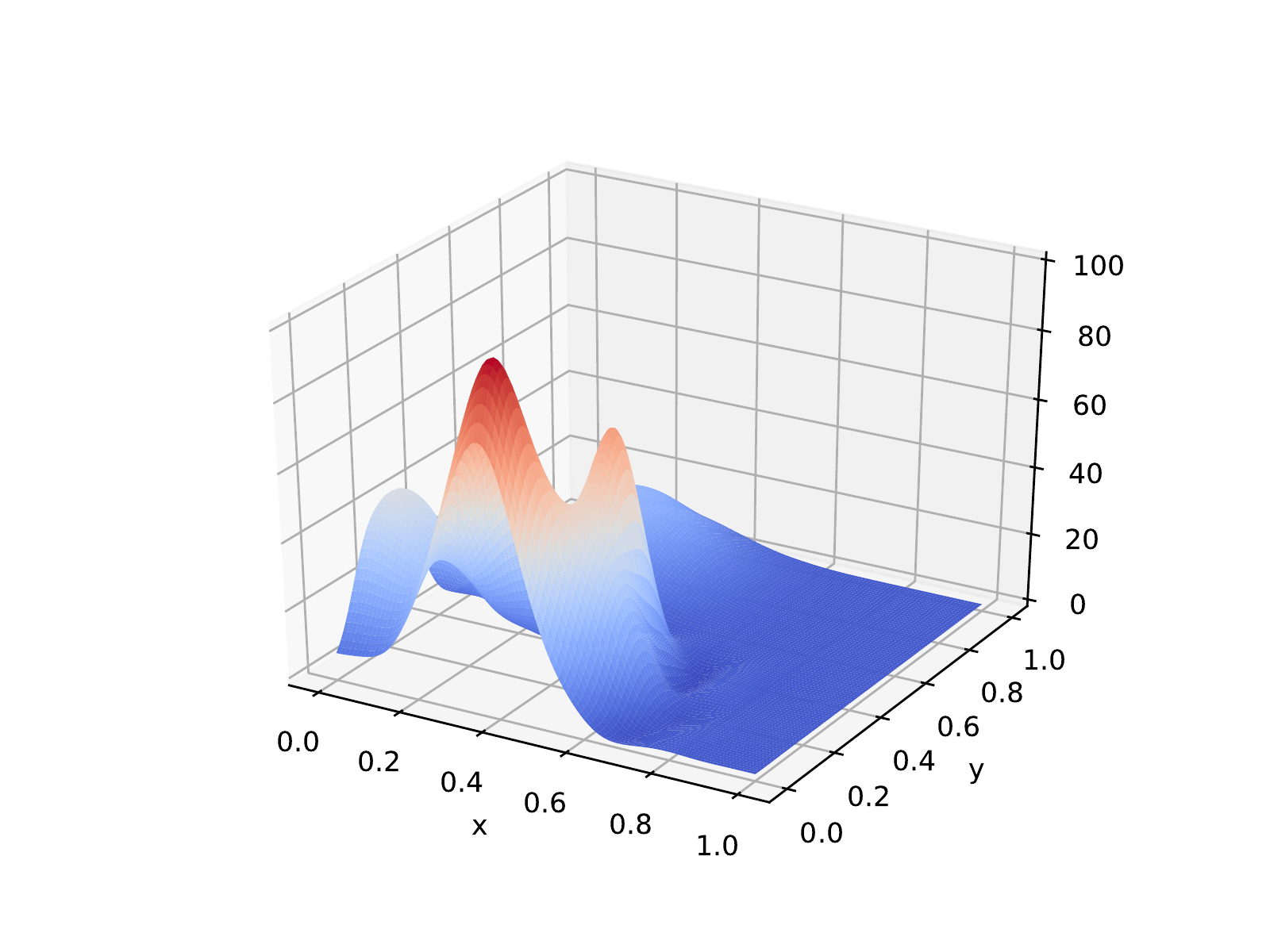}
  \caption{``True'' function used in instance \texttt{f5} (left), initial estimation (center) and estimation after probing (right).}
  \label{fig:f5}
\end{figure}

In order to visualize the improvements that are obtained by probing, we have prepared figures \ref{fig:f1} and \ref{fig:f5}, each with the true function, the initial estimation (in the cases shown, with 16 points), and with the estimation after probing with the orienteering tour.  As can be seen, the application of our method results in a clear improvement on the approximation of the true function.

\section{Conclusions}
\label{sec:conclusions}

This paper describes a problem arising in sea exploration, where the aim is to decide the schedule of a ship expedition for collecting information about the resources of the seafloor.  The setting involves the simultaneous use of tools from machine learning and combinatorial optimization.   We propose a method for its solution, dividing the process in three phases: assessment (building an indicator function that associates to each point in the sea a value for the interest in probing it),
planning (deciding on the position of points to probe in the next trip) and estimation (predicting a value of the resource level at any point on the surface).  The results obtained indicate that using the method here proposed clearly improves the quality of the estimation, by probing at promising points and adding the newly collected information to a regression step, which is based on Gaussian processes.

An interesting direction for future research is adapting the algorithm proposed here in order to deal with real-time changes in the path, reacting in the most appropriate way to information available as new observations arrive.
{ 
A realistic scenario for this is to commit to the next point to visit, but to leave the remainder part of the plan open to updates, after processing the information collected at that point.

Our algorithm has been proposed on the basis of realistic scenarios, but it has not been tested in practice; we hope the results presented provide a first step for its future adoption.
}

\subsubsection*{Acknowledgements}
This work was partially supported by project  "Coral - Sustainable Ocean Exploitation: Tools and Sensors/NORTE-01- 0145-FEDER-000036", financed by the North Portugal Regional Operational Programme (NORTE 2020), under the PORTUGAL 2020 Partnership Agreement, and through the European Regional Development Fund (ERDF).  We also thank Dr.\ Davi Pereira dos Santos for having checked and improved our code and text.

\bibliographystyle{spbasic}
\bibliography{coral}

\appendix

\section{Benchmark instances used}
\label{sec:instances}

\begin{figure}[!htbp]
  \centering
  \includegraphics[width=.33\textwidth,trim=100 30 50 50,clip=True]{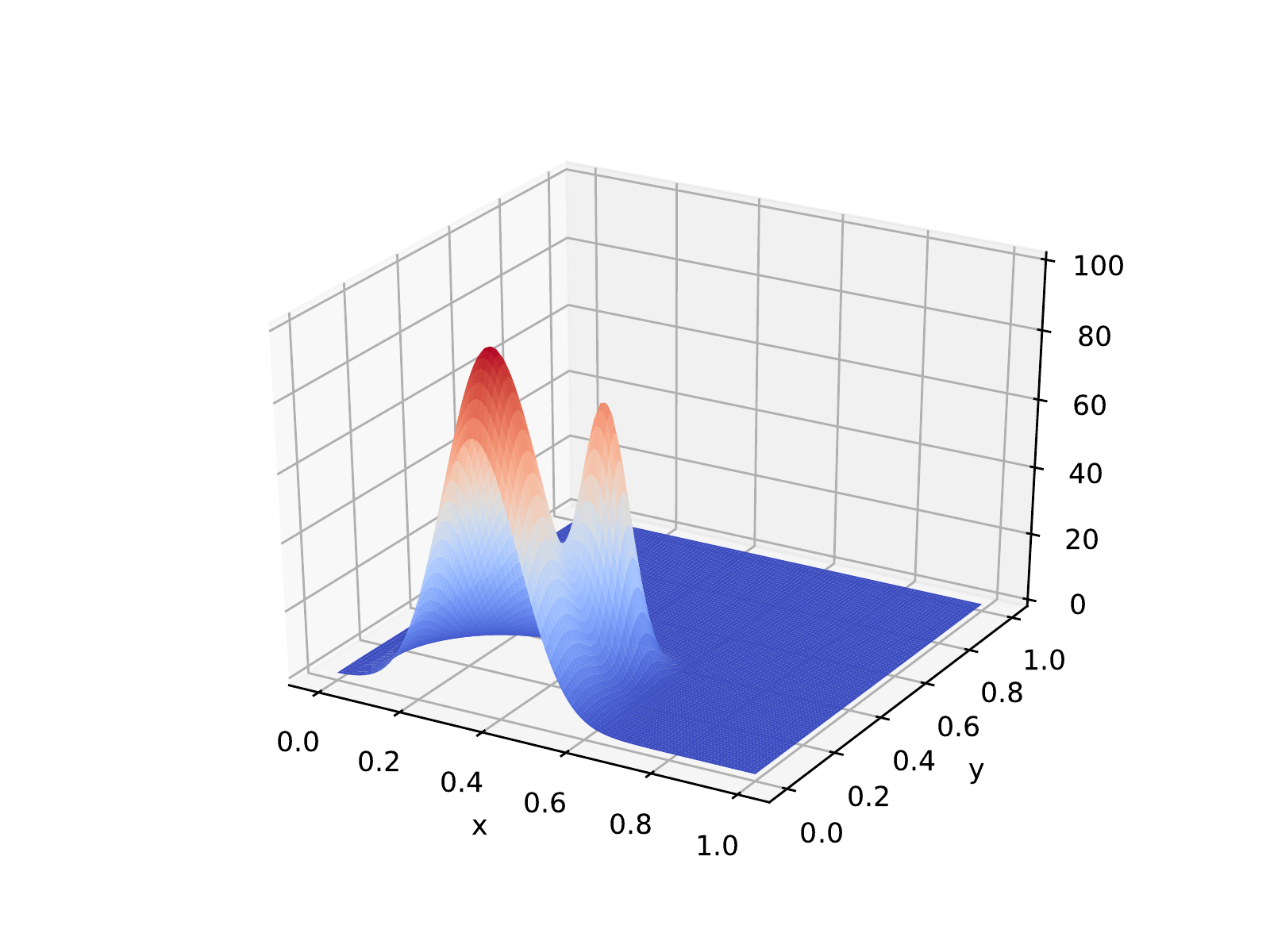}
  \includegraphics[width=.33\textwidth,trim=100 30 50 50,clip=True]{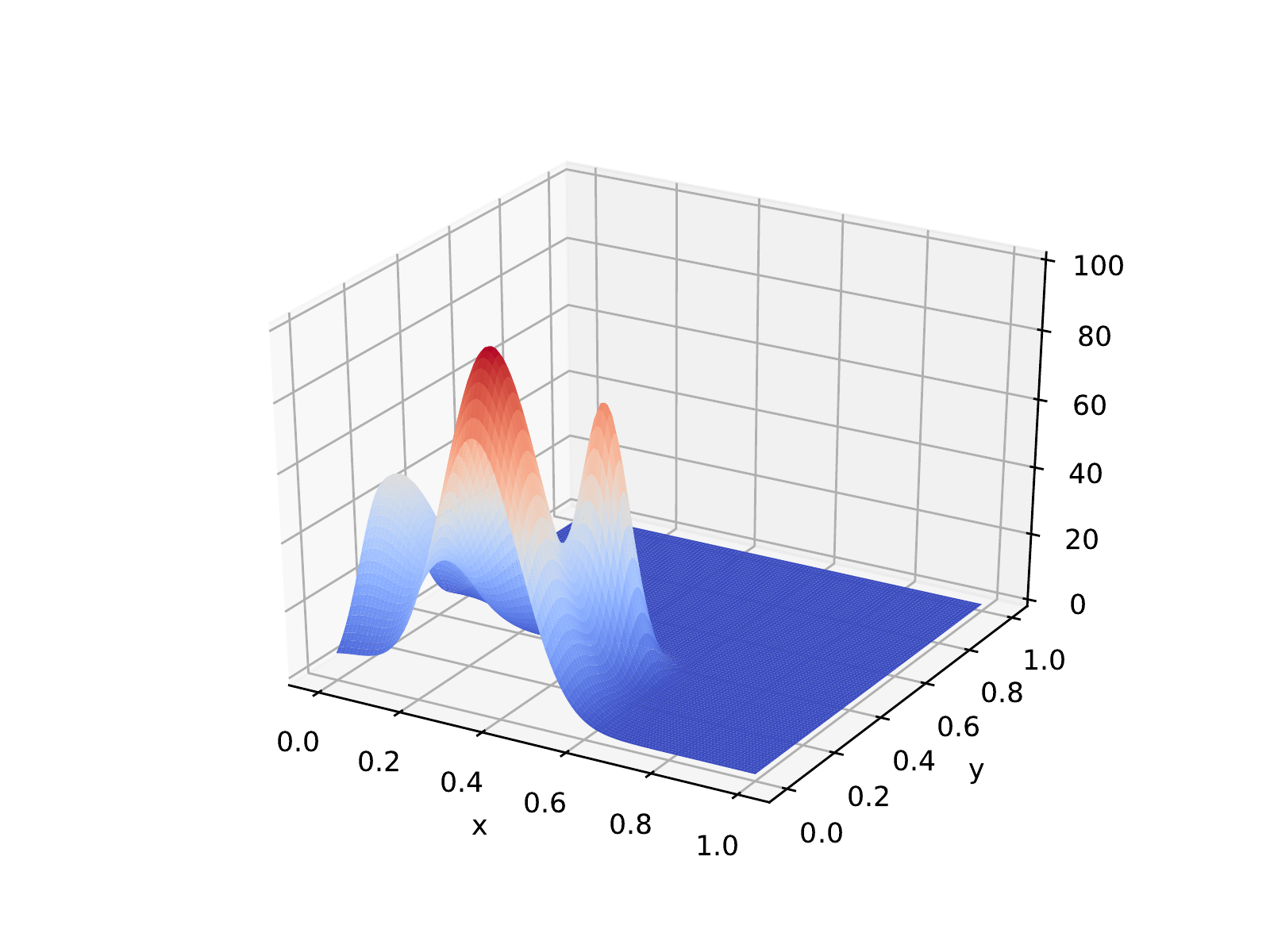} 
  \includegraphics[width=.33\textwidth,trim=100 30 50 50,clip=True]{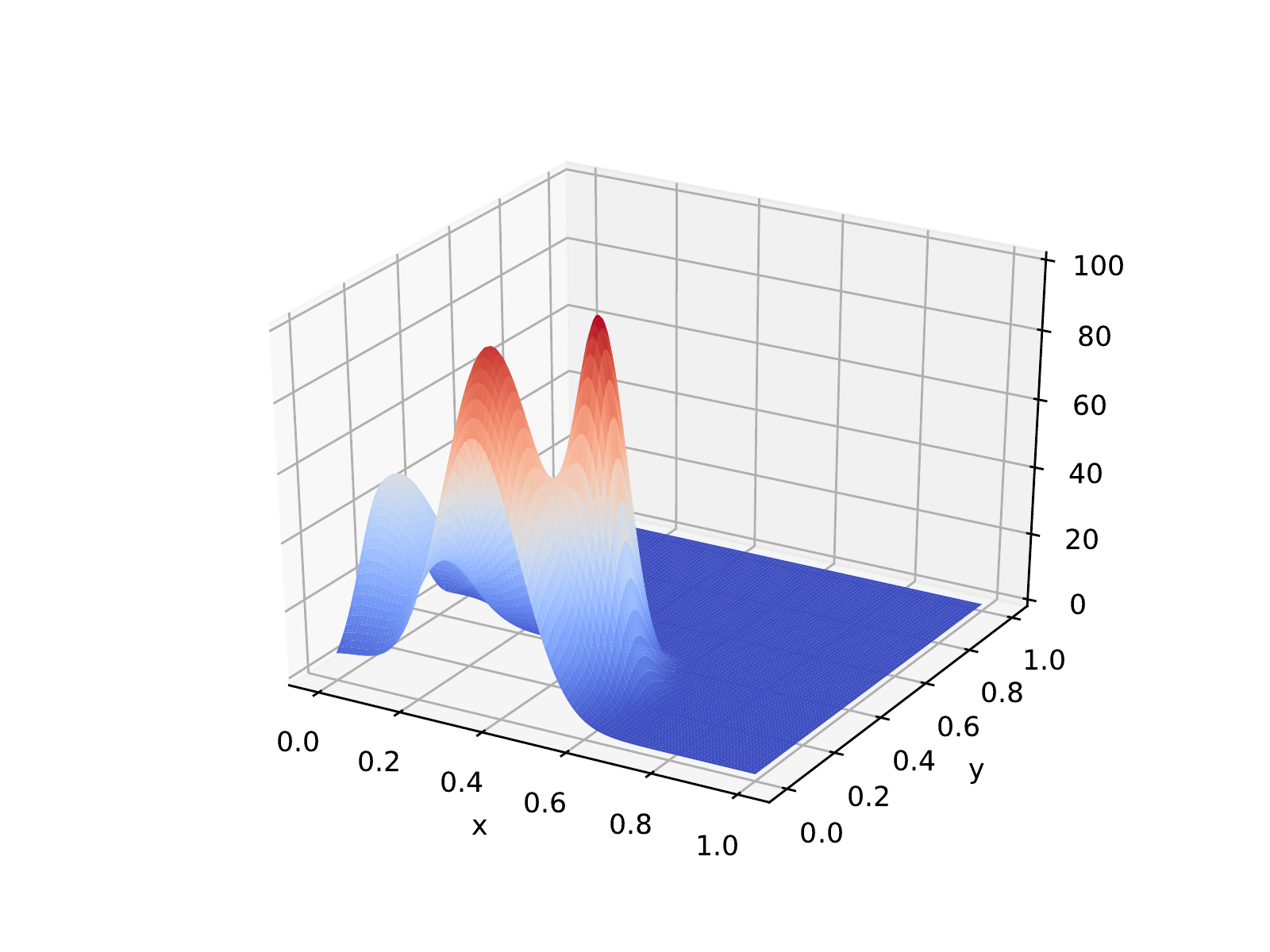} 
  \includegraphics[width=.33\textwidth,trim=100 30 50 50,clip=True]{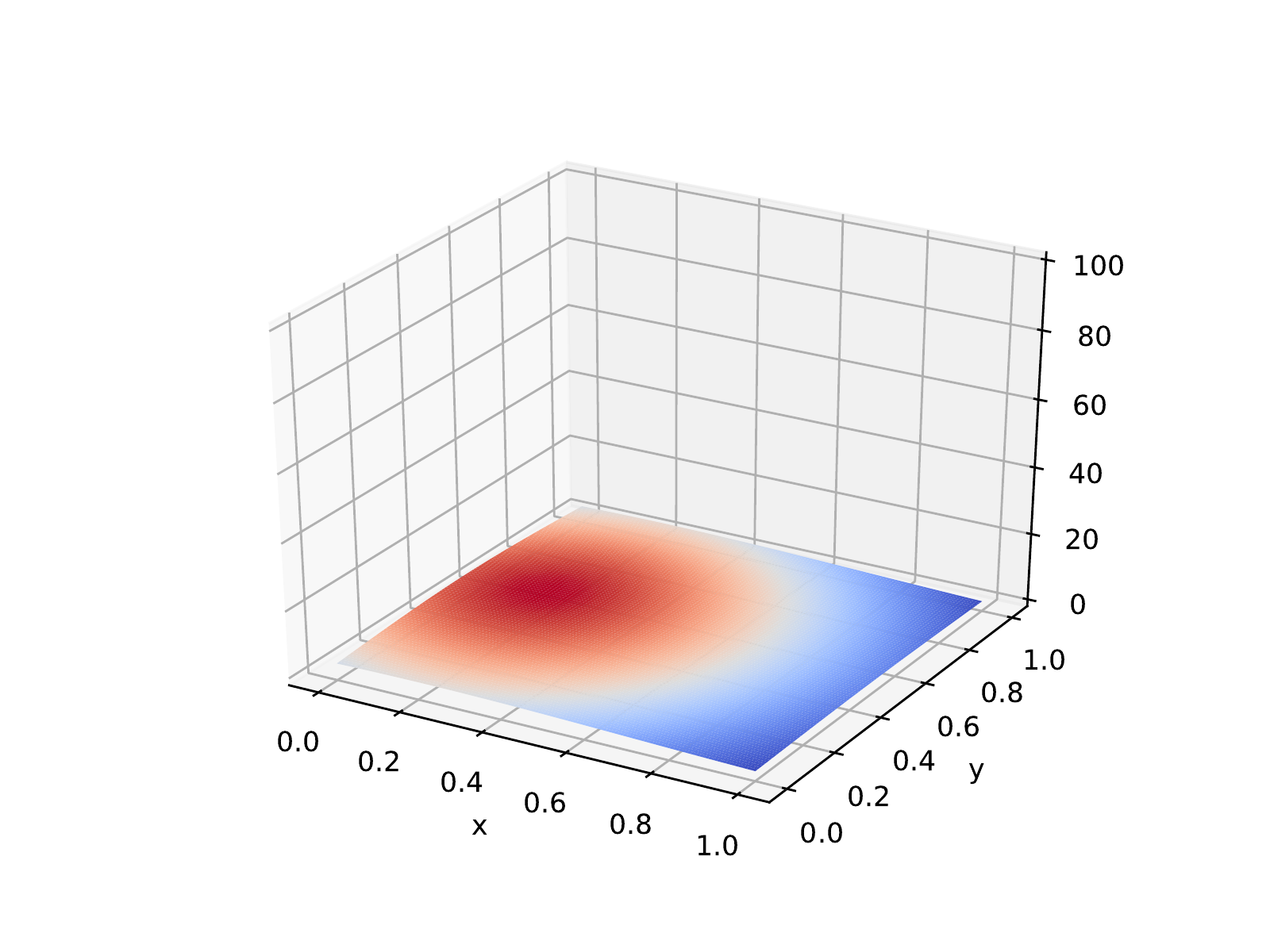}
  \caption{``True'' function used in instances \texttt{f2, f3} (top) and \texttt{f4, f6} (bottom).}
  \label{fig:fA}
\end{figure}

\begin{figure}[!hbtp]
  \centering
  \includegraphics[width=.33\textwidth,trim=100 30 50 50,clip=True]{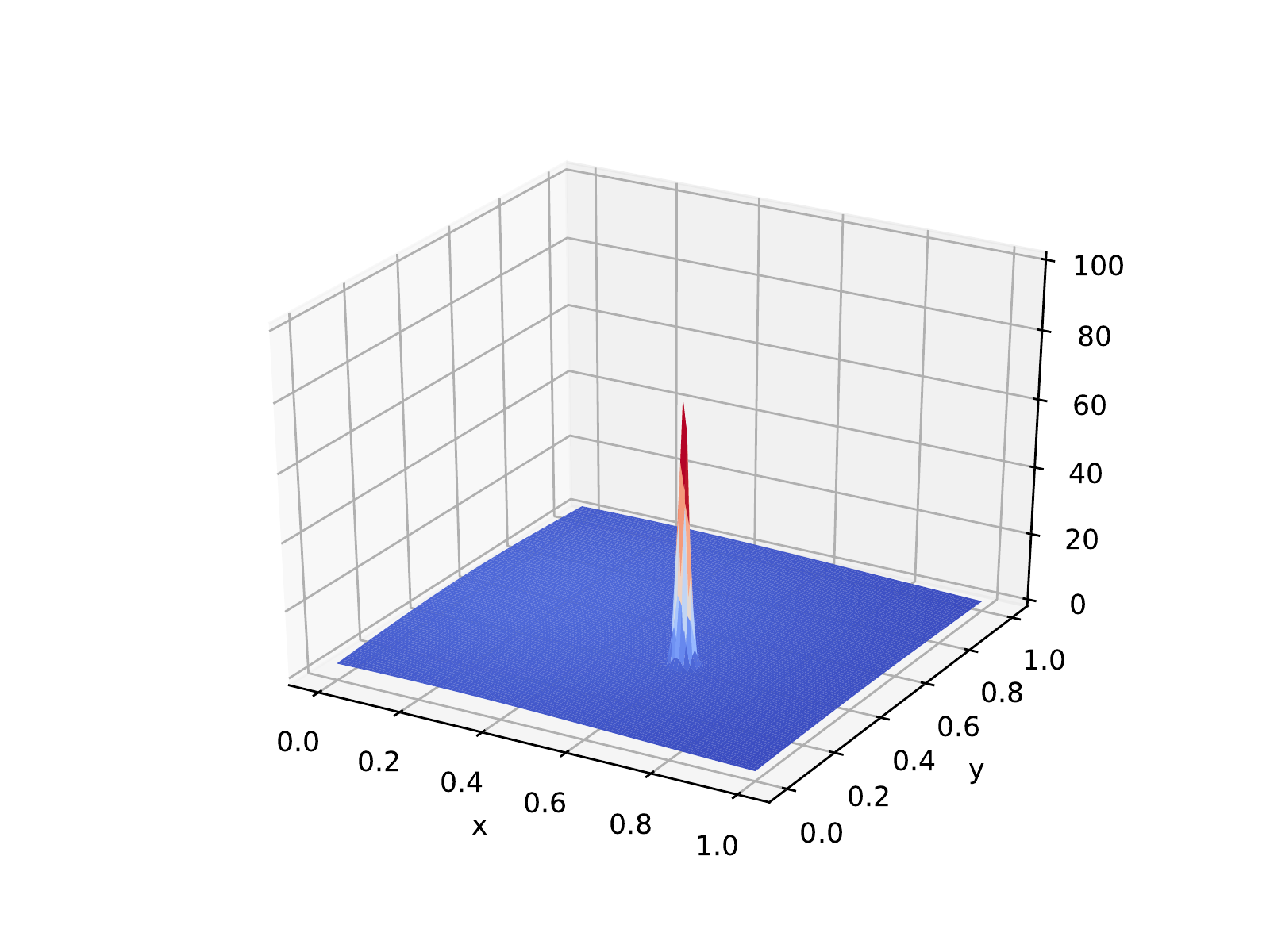}
  \includegraphics[width=.33\textwidth,trim=100 30 50 50,clip=True]{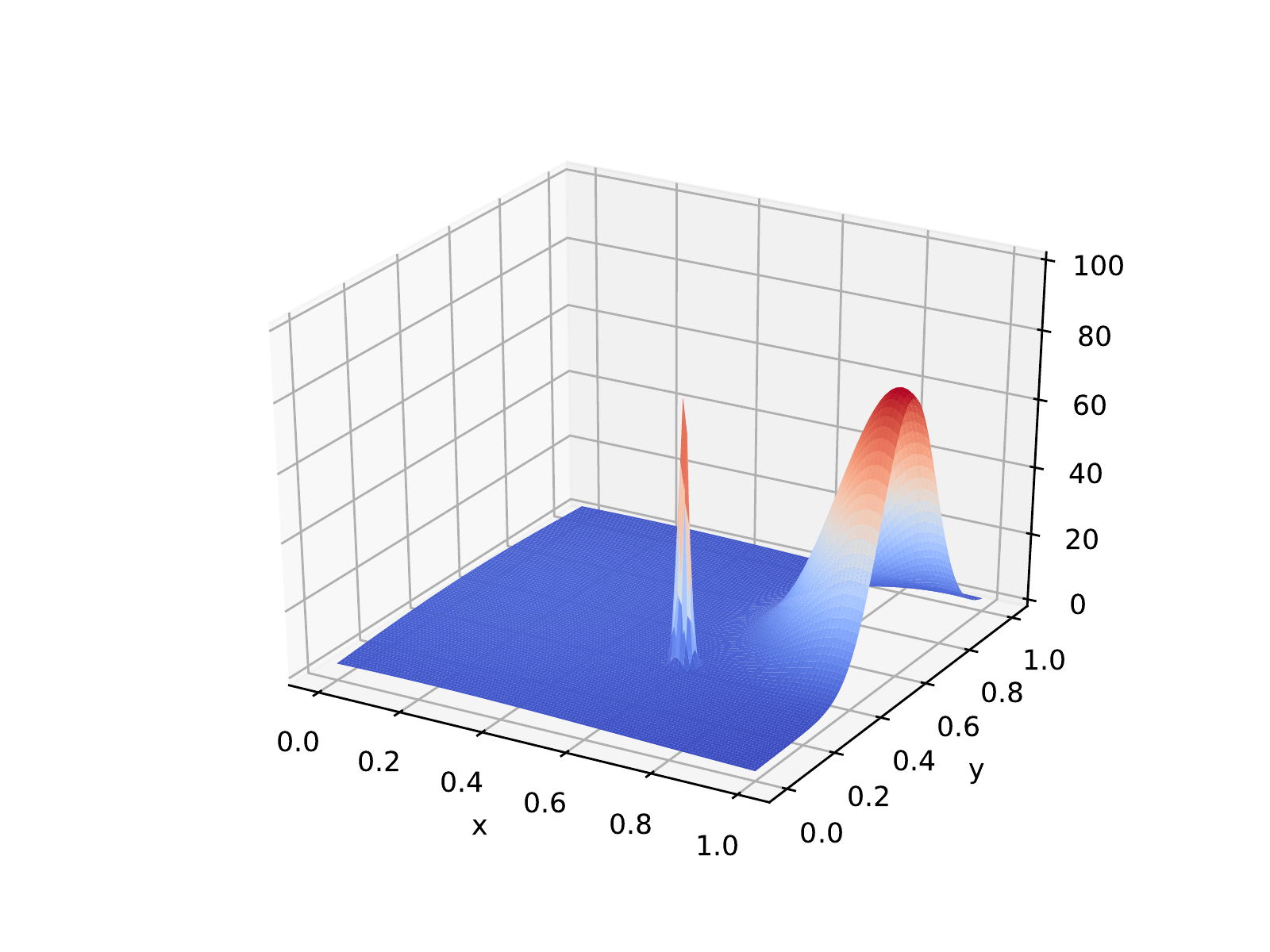}
  \includegraphics[width=.33\textwidth,trim=100 30 50 50,clip=True]{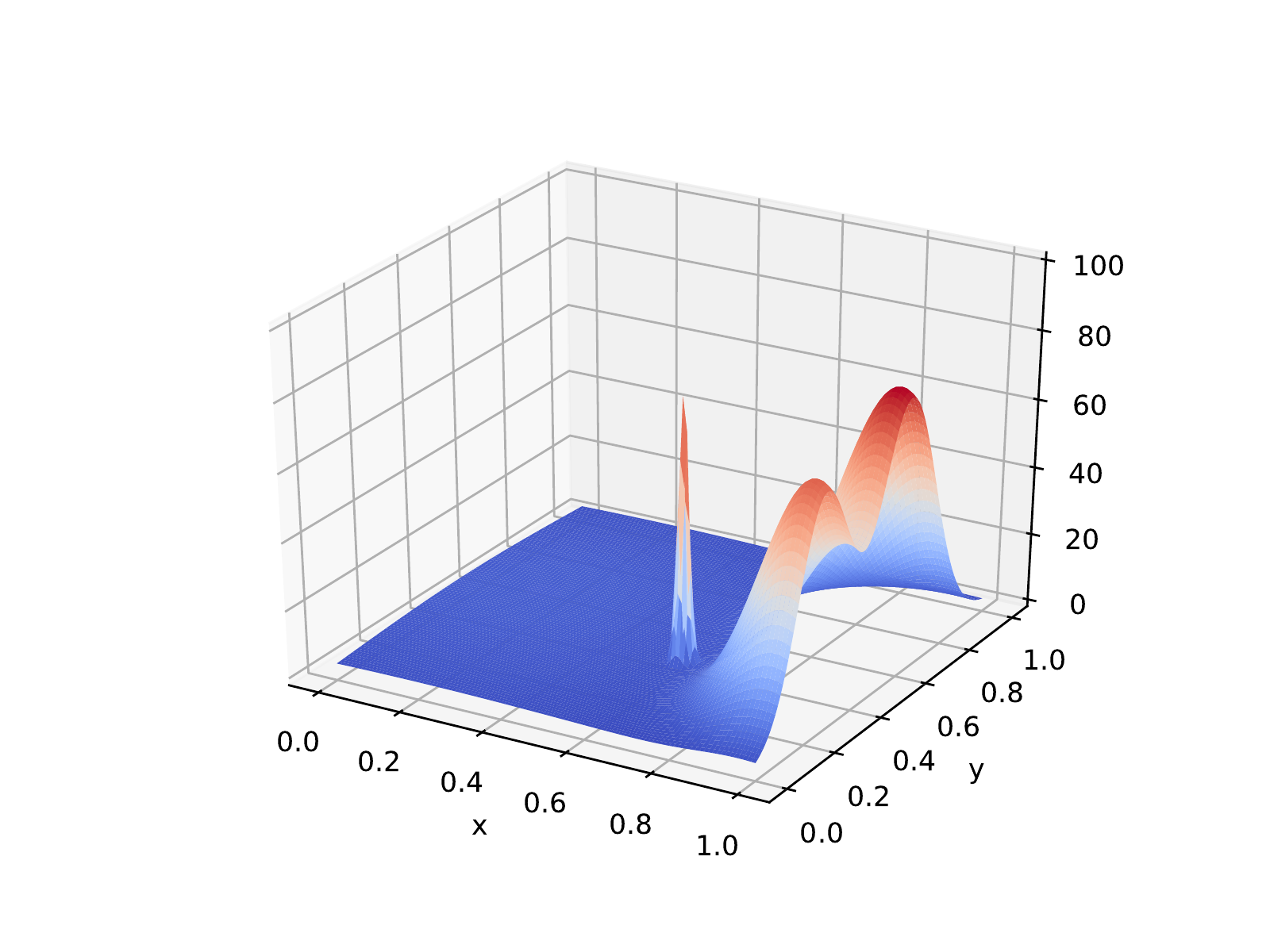}
  \includegraphics[width=.33\textwidth,trim=100 30 50 50,clip=True]{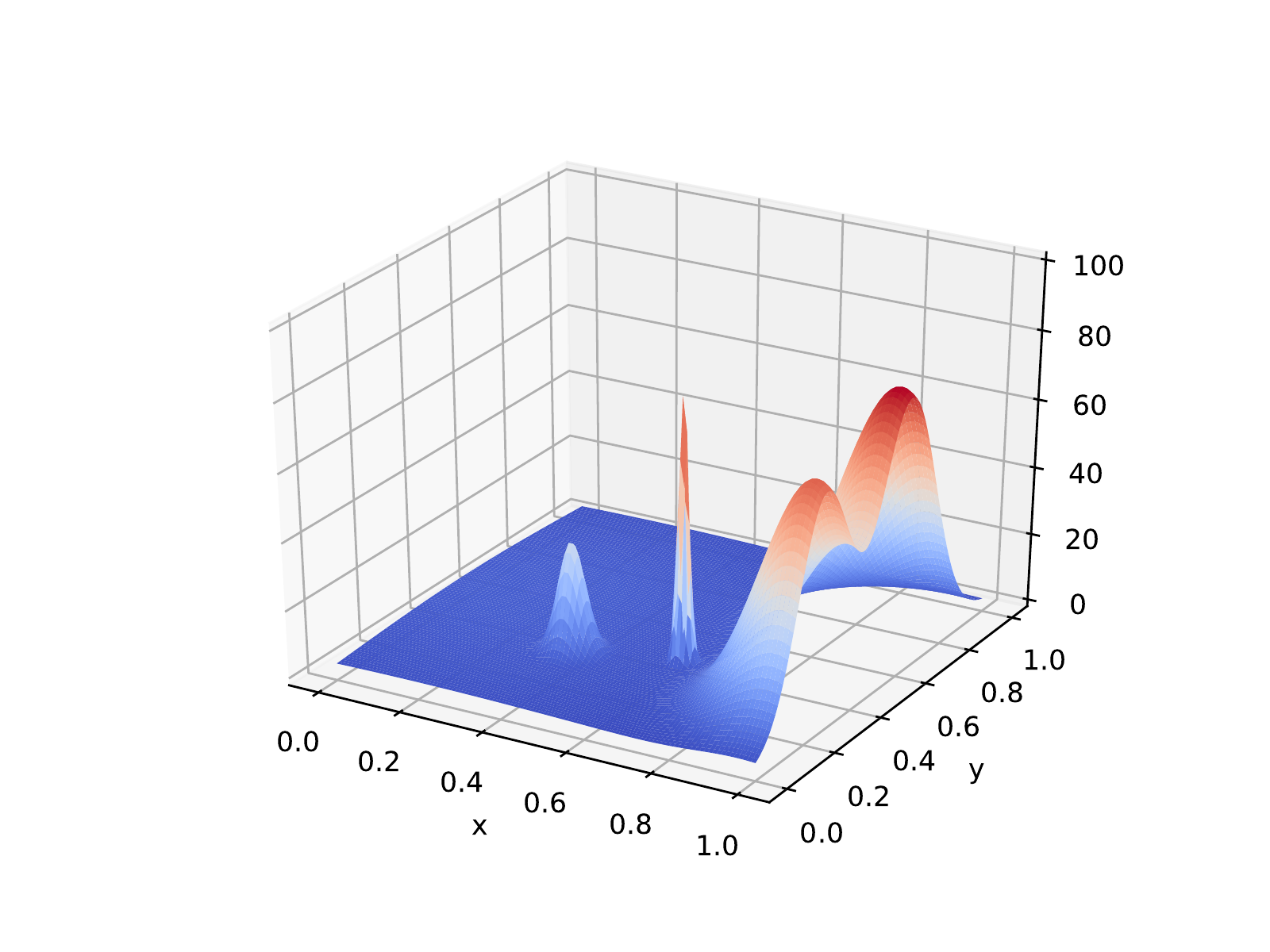}
  \caption{``True'' function used in instances \texttt{f7, f8} (top) and \texttt{f9, f10} (bottom).}
  \label{fig:fB}
\end{figure}

A complete description of the benchmark instances used is available as \emph{manuscript's data} and at the author's homepage\footnote{\url{http://www.dcc.fc.up.pt/~jpp/code/CORAL}}.
In summary, there are 10 different ``true'' functions that should be guessed, \texttt{f1} to \texttt{f10}; an illustration of their shapes is provided in figures \ref{fig:f1} to~\ref{fig:fB}.
{ 
  Functions \texttt{f1} to \texttt{f5} have increasing complexity, are are provided for the purpose of algorithm tuning.  Functions \texttt{f6} to \texttt{f10} are also increasingly complex; \texttt{f6} is very smooth and should pose no problems to the fit, while \texttt{f7} has a steep maximum that is very unlikely to be found with a small number of function evaluations.
}

The area $S$ being studied is considered to be $(x,y) \in [0,1] \times [0,1]$.  The set of points initially available are noiseless evaluations of functions \texttt{f1} to \texttt{f10}, either in a regular grid (\eg, as in Table~\ref{tab:f1-16grid}) or randomly spread in $S$ (as in Table~\ref{tab:f1-16rand}).

Other relevant data are: the probing time $t=1$ time units; the speed of traveling $s=1$; and parameter $T=100$.  The starting and ending nodes are located at coordinates (0,0).    

As for the evaluation of the quality of the algorithms, the differences between the true and the predicted values, $w(x,y)$ and $v(x,y)$,  are assessed on $K$ points $E$  on a mesh $x,y \in \{0, 0.01, \ldots, 1\}$.

\begin{table}[!htbp]
  \centering
  \caption{Evaluations of f1 on a grid.}
\begin{tabular}{l|rrrr}
\diagbox{$\bar{x}$}{$\bar{y}$}
      &  0.20 &  0.40 &  0.60 &  0.80\\\hline
  0.2 &  0.00 &  0.02 &  8.21 & 60.65\\
  0.4 &  0.00 &  0.00 &  0.15 &  1.11\\
  0.6 &  0.00 &  0.00 &  0.00 &  0.00\\
  0.8 &  0.00 &  0.00 &  0.00 &  0.00\\
\end{tabular}
  \label{tab:f1-16grid}
\end{table}

\begin{table}[!htbp]
  \centering
  \caption{Evaluations of f1 at random positions.}
  \label{tab:f1-16rand}
  \sisetup{
    round-mode = places,
    round-precision = 3 }%
  \begin{tabular}{SSS}
    {$x$} & {$y$} & {$f(x,y)$} \\\hline
0.0254458609934608	& .5414124727934966	& 0.0 \\ 
0.029040787574867943	& .22169166627303505	& 0.17910654005215518\\
0.0938595867742349	& .02834747652200631	& 3.338438486158641\\
0.13436424411240122	& .8474337369372327	& 0.0 \\ 
0.21659939713061338	& .4221165755827173	& 0.084613952016198\\
0.22876222127045265	& .9452706955539223	& 0.0 \\ 
0.23308445025757263	& .2308665415409843	& 13.845273128139127\\
0.43788759365057206	& .49581224138185065	& 0.008218117557316357\\
0.4453871940548014	& .7215400323407826	& 0.0 \\ 
0.49543508709194095	& .4494910647887381	& 0.027187155905827275\\
0.651592972722763	& .7887233511355132	& 0.0 \\ 
0.762280082457942	& .0021060533511106927	& 0.016852780725354247\\
0.763774618976614	& .2550690257394217	& 0.003676848494094813\\
0.8357651039198697	& .43276706790505337	& 0.0 \\ 
0.9014274576114836	& .030589983033553536	& 0.0 \\ 
0.9391491627785106	& .38120423768821243	& 0.0 \\ 
  \end{tabular}
\end{table}

\end{document}